\documentclass[10pt,twocolumn,letterpaper]{article}

\usepackage{iccv}
\usepackage{times}
\usepackage{epsfig}
\usepackage{graphicx}
\usepackage{amsmath}
\usepackage{amssymb}
\usepackage{algorithm2e}
\usepackage{algorithmicx}
\usepackage{bm}
\usepackage{epstopdf}
\usepackage{color}

\definecolor{red}{rgb}{1,0,0}

\usepackage[pagebackref=true,breaklinks=true,letterpaper=true,colorlinks,bookmarks=false]{hyperref}

\iccvfinalcopy % *** Uncomment this line for the final submission

\def\iccvPaperID{230} % *** Enter the ICCV Paper ID here

\DeclareMathOperator*{\argmin}{arg\,min}

\newcommand{\mbr}[1]{\mathrm{\bm{\mathbf{#1}}}}
\newcommand{\mr}[1]{\mathrm{#1}} 

\ificcvfinal\fi
\begin{document}

%%%%%%%%% TITLE
\title{A Picture is Worth a Billion Bits: Real-Time Image Reconstruction from Dense Binary Pixels}

\author{Tal Remez$^*$\\
{\tt\small talremez@mail.tau.ac.il}
\and
Or Litany$^*$\\
{\tt\small orlitany@post.tau.ac.il}\\
Tel-Aviv University\\
\tiny{$*$ Equal contributors}\\
\and
Alex Bronstein\\
{\tt\small bron@eng.tau.ac.il}
}

\maketitle
%\thispagestyle{empty}

%%%%%%%%% ABSTRACT
\begin{abstract}
%The ABSTRACT is to be in fully-justified italicized text, at the top
%   of the left-hand column, below the author and affiliation
%   information. Use the word ``Abstract'' as the title, in 12-point
%   Times, boldface type, centered relative to the column, initially
%   capitalized. The abstract is to be in 10-point, single-spaced type.
%   Leave two blank lines after the Abstract, then begin the main text.
%   Look at previous ICCP abstracts to get a feel for style and length.

The pursuit of smaller pixel sizes at ever increasing resolution in digital image sensors is mainly driven by the stringent price and form-factor requirements of sensors and optics in the cellular phone market. Recently, Eric Fossum proposed a novel concept of an image sensor with dense sub-diffraction limit one-bit pixels (\emph{jots}) \cite{fossum2005sub}, which can be considered a digital emulation of silver halide photographic film. This idea has been recently embodied as the EPFL Gigavision camera.
A major bottleneck in the design of such sensors is the image reconstruction process, producing a continuous high dynamic range image from oversampled binary measurements. The extreme quantization of the Poisson statistics is incompatible with the assumptions of most standard image processing and enhancement frameworks. The recently proposed maximum-likelihood (ML) approach addresses this difficulty, but suffers from image artifacts and has impractically high computational complexity.
In this work, we study a variant of a sensor with binary threshold pixels and propose a reconstruction algorithm combining an ML data fitting term with a sparse synthesis prior. We also show an efficient hardware-friendly real-time approximation of this inverse operator.
Promising results are shown on synthetic data as well as on HDR data emulated using multiple exposures of a regular CMOS sensor.

%\small\begin{verbatim}
%    \usepackage[dvips]{graphicx} ...
%    \includegraphics[width=0.8\linewidth]
%                   {./figures/pipeline.jpg}
%\end{verbatim}

\end{abstract}

%%%%%%%%% BODY TEXT
\section{Introduction}
\begin{figure*}[th!]
  \centering
  \includegraphics[width=1\linewidth]{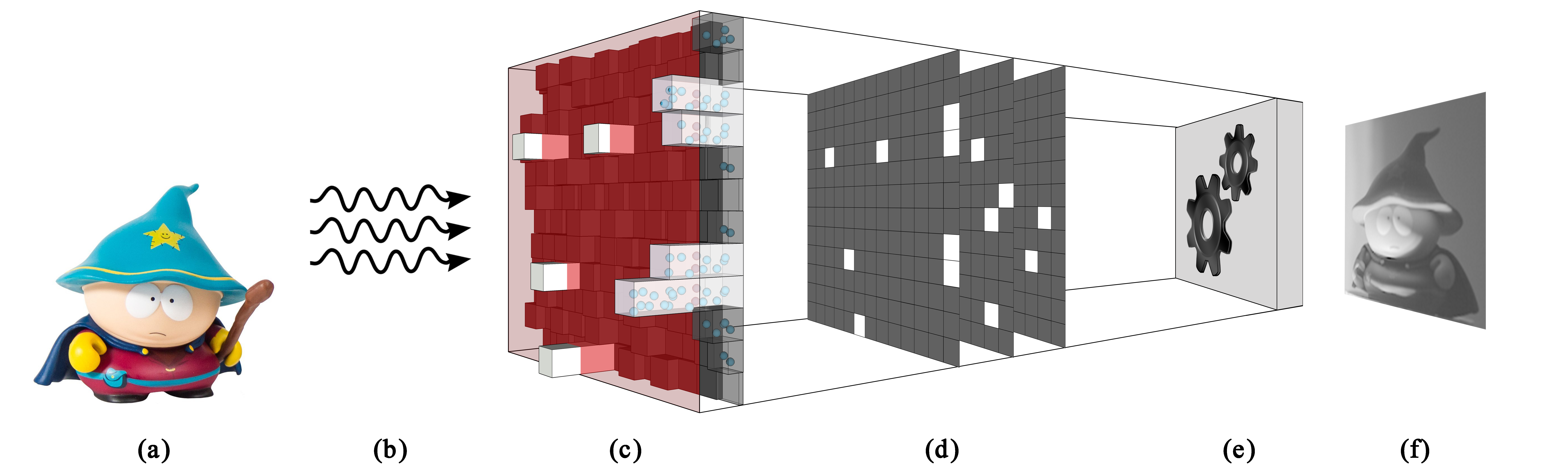}
  \caption{\small \textbf{Image acquisition and reconstruction pipeline.}
                          Presented is the binary image acquisition and reconstruction process.
                          (a) the scene being photographed; (b) the photon flux reaching the binary sensor; (c) the sensor collecting the photons and thresholding the photoelectron count; (d) binary images produced by the sensor; (e) our reconstruction algorithm; (f) the reconstructed image.
}
  \label{pipeline}
\end{figure*}

The resolution of image sensors has been steadily increasing in the past two decades, more or less following Moore's law. Simultaneously, the advent of the cell phone market has placed stringent constraints on the form factor and price of sensors and optical components, creating very strong incentives for camera miniaturization. These two trends are driving the pursuit of smaller and more densely packed pixels. Today, mainstream CMOS sensor technology allows to manufacture pixels in the $1$ micron range, and much smaller pixel sizes are feasible.

However, such a miniaturization comes at the price of pixel quality. Small pixels have an inherently limited full-well capacity and consequently have low SNR and a poor dynamic range. This is already a significant obstacle in mainstream CMOS sensors showing poor capabilities in high dynamic range (HDR) and low light scenes. Furthermore, the resolution of miniature sensors falls below the diffraction limit of existing optics (for comparison, the Airy disk diameter of a lens with $f/2$ at $555$ nm is about $2.7$ $\mu$m).

In \cite{fossum2005sub,nakamura2005image}, Eric Fossum proposed the novel concept of “digital film sensor” composed of sub-micron threshold pixels (\emph{jots}) producing binary outputs. A dense array with billions of jots resembles the traditional photographic film in which silver halide crystals of comparable size form light-sensitive grains, or the human eye in which the role of the binary threshold pixels is played by the cones and rods. When a photon hits a grain, it results in the liberation of a silver atom that in the development process ``exposes" the entire grain. Detailed probability analysis shows that the density of the exposed grains depends non-linearly on the illumination intensity, effectively compressing high dynamic range into a small interval. The response of photoreceptor firing rates in our eyes exhibits similar behaviour.
Thus, if made sufficiently low noise, dense binary pixels can mimic the high dynamic range of photographic films or animal visual systems. A proof of this concept has been shown in the EPFL Gigavision camera \cite{yang2009image} with $0.75$ $\mu$m binary pixels.

While binary sensors seem to be a promising "out-of-the-box" solution for HDR imaging, their output is not directly usable and requires an image reconstruction process. Very low photon count in each individual pixel demands accurate modelling of the underlying Poisson statistics, which renders many mainstream image reconstruction and enhancement approaches inapplicable. Moreover, even the recent low light image enhancement algorithms such as \cite{giryes2013sparsity, toEorNotToE} based on careful modelling of the noise are not directly applicable due to the extreme quantization of the binarized signal.

In \cite{yang2012bits}, a maximum-likelihood approach to image reconstruction from binary pixels was proposed. While the algorithm is based on the minimization of a convex objective, its computational complexity is prohibitive for any time-critical application.

Signal reconstruction from binary measurements has been recently studied in the compressed sensing community \cite{jacques2011robust}. However, the authors used the standard Euclidean data fitting term unsuitable for Poisson statistics.
Sparse representation techniques have also been applied to the closely related problem of \emph{inverse halftoning}, In \cite{son2012inverse} it was proposed to co-train a pair of dictionaries, one suited for representing halftoned images, and the other for the corresponding continues image while enforcing the same representation in both dictionaries. A reconstruction of the continuous valued patch is achieved by plugging the binary patch's representation in its dictionary into the corresponding continuous dictionary. This approach lacks the physical model of the binarization procedure, which is essential in our case. In \cite{mairal2012task} it was demonstrated that task-specific dictionary learning suited for the non-linear inverse problem yields state-of-the-art inverse halftoning results. While, the use of the Euclidean data fitting term and the specific Floyd-Steinberg \cite{floyd1976adaptive} halftoning operator is unsuitable for binary imaging, as we detail in the sequel, the present work follows a similar task-driven learning spirit.

\textbf{Contributions}
In this paper, we consider a variant of a binary thresholded pixel array, for which we formulate the image reconstruction problem as the minimization of a convex objective combining a likelihood term similar to \cite{yang2012bits} and a patch-based sparse synthesis prior. Our objective function can be viewed as a variant of the objective used in \cite{giryes2013sparsity, toEorNotToE} for low light image denoising with a different data fitting term correctly capturing the signal formation model, or as a regularised version of the ML estimator from \cite{yang2012bits}.

We simulate the binary image acquisition process using multiple exposures of a regular Canon DSLR CMOS sensor and show that our approach outperforms the previously proposed ML estimator in terms of image quality. However, since the minimization requires an iterative algorithm, the resulting computational complexity is still prohibitive. To address this bottleneck, we show how to learn a fixed complexity and latency approximation of the image reconstruction operator inspired by the recent works \cite{gregor-icml-10,sprechmann2012learning}. We show that comparable reconstruction quality is achieved at a fraction of the computational complexity of the iterative algorithm. To the best of our knowledge, no efficient signal recovery algorithm exist for non-Euclidian fitting terms.

The rest of the paper is organized as follows: In Section \ref{sec_forward} we describe the acquisition process using a binary sensor, in Section \ref{sec_MLSP} we formulate the reconstruction model, in Section \ref{sec_fastApprox} we show a fast approximation to the reconstruction algorithm. Section \ref{sec_results} is dedicated to experimental evaluation of our algorithm. Finally, Section \ref{sec_conclusions} concludes the paper.

\section{Binary image formation model}
\label{sec_forward}
We denote by $\mbr{x}$ the radiant exposure at the camera aperture measured over a given time interval.
This exposure is subsequently degraded by the optical point spread function denoted by the operator $\mbr{H}$, producing the exposure on the sensor $\mbr{\lambda} = \mbr{Hx}$.
The number of photoelectrons $e_{jk}$ generated at pixel $j$ in time frame $k$ follows the Poisson distribution with the rate $\lambda_j$,
\vspace{-2 mm}
\begin{equation}\label{}
    \mathrm{P}\left( e_{jk} = n \mid \lambda_j \right) = \frac{{e^{ - \lambda_j } \lambda_j^n }}{{n!}}
\end{equation}
A binary pixel compares the accumulated charge against a pre-determined threshold $q_j$\footnote{We assume $q_j$ to be fixed in time. Various CMOS designs with time-varying pixel threshold are possible.}, outputting a one-bit measurement $b_{jk}$ (see Figure \ref{pipeline} for an illustration). Thus, the probability of a single binary pixel $j$ to assume an "off" value in frame $k$ is
\vspace{-1 mm}
\begin{equation}\label{}
    \begin{split}
    p_{j} &= \mathrm{P}(b_{jk}=0 \mid q_j,\lambda_j )=\mathrm{P}(e_{jk} < {q_j} \mid q_j,\lambda_j) %= \sum_{i=0}^{q_j-1}{\frac{e^{-\lambda_j}\lambda_j^i}{i!}} ,
    \end{split}
\end{equation}

\noindent therefore we can write
\begin{equation}\label{}
\mathrm{P}(b_{jk} \mid q_j,\lambda_j) = (1-b_{jk})p_{j} + b_{jk}(1-p_{j}) .
\end{equation}
 Assuming independent measurements, the negative log-likelihood of the radiant exposure $\mbr{x}$ given the measurements $b_{jk}$ is given by
\begin{equation}\label{likelihood}
\ell(\mbr{x} \mid \mbr{B}) = \mathrm{const} - \sum_{j,k} \log \mathrm{P}(b_{jk} \mid q_j, \lambda_j),
\end{equation}
In \cite{EPFL-REPORT-166345}, a maximum-likelihood approach was proposed for solving \ref{likelihood}.

\section{Maximum-likelihood with reconstruction sparse prior}
\label{sec_MLSP}
Standard optimization techniques were used in ~\cite{EPFL-REPORT-166345} to solve problem \ref{likelihood}. In all our experiments problem \ref{likelihood} was solved on the entire image using the trust-region-reflective algorithm from MATLAB optimization toolbox. Since this approach assumes no prior, it requires a large number $K$ of binary measurements in order to achieve a good reconstruction (see Figure \ref{MultipleExpGraph}). Sparsity priors have been shown to give state-of-the-art results in denoising tasks in general, and particularly in low light settings where Poisson noise statistics become dominant \cite{giryes2013sparsity,toEorNotToE}. In the following, we show that by introducing a spatial prior over the reconstructed image patches one may significantly reduce the number of measurements required without hampering the reconstruction quality. Assuming that the radiant exposure $\mbr{\lambda}$ can be expressed by a kernelized sparse representation $\mbr{\lambda} = \mbr{H}\rho(\mbr{Dz})$, where $\mbr{D}$ is a dictionary (e.g. globally trained via k-SVD \cite{aharon2006svd}) and $\rho$ is an element-wise intensity transformation function, we may express the estimator as $\hat{\mbr{x}} = \rho(\mbr{D} \hat{\mbr{z}})$, where
\begin{equation}\label{problemFormulationWithSaprsity}
    \mbr{\hat{z}} = \argmin_{\mbr{z}}{\ell(\rho(\mbr{Dz}) \mid \mbr{B}) + \mu \|\mbr{z}\|_1} ,
\end{equation}
and $\|\mbr{z}\|_1$ denotes the sparsity encouraging $\ell_1$ norm of the coefficient vector $\mbr{z}$. In all the experiments presented in this work, we chose $\rho$ to be the hybrid exponential-linear function introduced in \cite{toEorNotToE},
\begin{equation}
\label{rho}
\rho(x) =
\left\{
    \begin{array}{ll}
        c\exp(x) & x \leq 0 \\
        c(1+x) & x > 0
    \end{array}
\right .
\end{equation}
that was shown to have several advantages for Poisson image reconstruction as it enforces image non negativity constraints, while presenting relatively low Lipschitz constants across all intensity levels.

Since the negative log-likelihood data fitting term is convex (see supplementary material for a proof) with a Lipschitz-continuous gradient, problem (\ref{problemFormulationWithSaprsity}) can be solved using proximal algorithms such as the iterative shrinkage thresholding algorithm (ISTA), its accelerated version \cite{daubechies2004iterative} or FISTA \cite{beck2009fast}.
For completeness, the FISTA algorithm is summarized as Algorithm \ref{algo_fista}, defining $\sigma_\theta$ as the coordinate-wise shrinking function with threshold $\theta$ and $\eta$ as the step size. The gradient of the negative log-likelihood computed at each iteration is given by
\vspace{-1 mm}
\begin{equation}
\frac{\partial\ell}{\partial \mbr{z}} = \mbr{D}^T \rm{diag}(\rho'(\mbr{D}\mbr{z}))\mbr{H}^T \nabla \ell(\mbr{H}\rho(\mbr{D}\mbr{z}) | \mbr{B} ) .
\end{equation}

\begin{algorithm}[h!]
\KwIn{Binary measurements $\mbr{B}$, step size $\eta$}
\KwOut{Reconstructed image $\hat{\mbr{x}}$ }
initialize $\mbr{z}^* = \mbr{z} = 0$, $\beta < 1$, $m_0 = 1$    \\
\For{$t=1,2,...,\mbox{until convergence}$}{

     \textit{//Backtracking} \\
     \While{$\footnotesize{\ell(\mbr{z}^*) \geq \ell(\mbr{z}) + <\mbr{z}^* - \mbr{z},\frac{\partial\ell}{\partial \mbr{z}}> + \frac{1}{2\eta} \| \mbr{z}^* - \mbr{z} \|_2^2 }$}{
     $\eta = \eta\beta$\\
     $ \mbr{z}^* = \sigma_{\mu\eta}(\mbr{z} - \eta\frac{\partial\ell}{\partial \mbr{z}}) $
     }
     \textit{//Step} \\
     \vspace{-1 mm}
     $m_{t+1} = \frac{1+\sqrt{1+4m_t^2}}{2}$\\
     $\mbr{z} = \mbr{z}^* + \frac{m_t-1}{m_{t+1}}(\mbr{z}^* - \mbr{z}) $\\

 }
 $\hat{\mbr{x}}=\rho(\mbr{D}\mbr{z})$
 \vspace{2 mm}
 \caption{FISTA with backtracking.}
\label{algo_fista}
\end{algorithm}
\noindent If the Lipschitz constant $L$ of $\frac{\partial \ell}{\partial \mbr{z}}$ or an upper bound thereof is easy to compute, backtracking can be replaced by a fixed step size $\eta < 1/L$.  The algorithm is reduced to the less efficient ISTA by fixing $m_t = 0$.

\section{Fast approximate reconstruction}
\label{sec_fastApprox}
\begin{figure}[th]
  \centering
  \includegraphics[width=1\columnwidth]{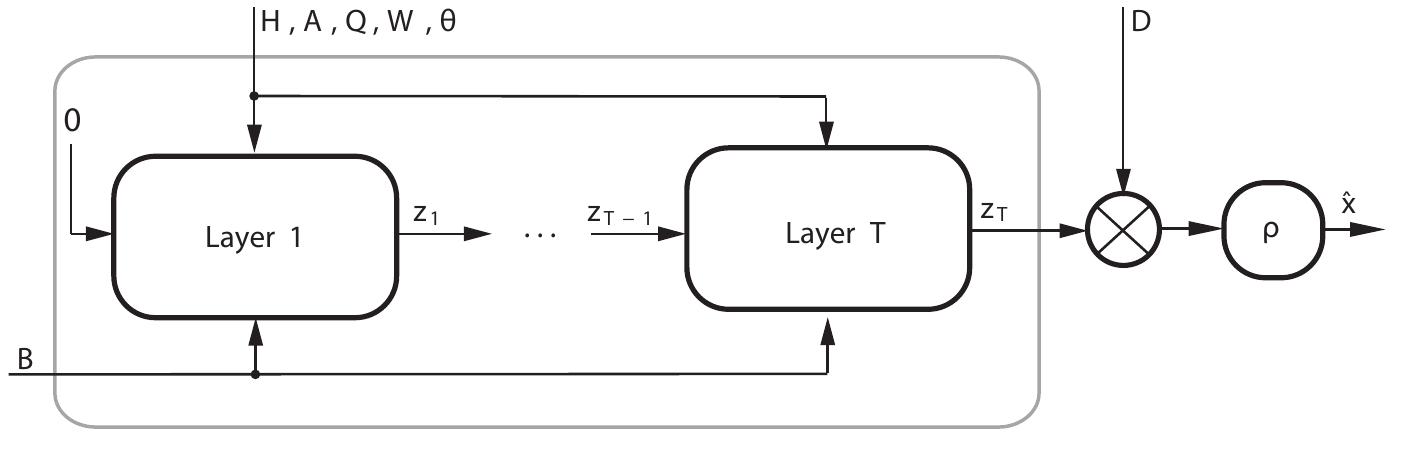}
%  $\mbr{z}_0  \mbr{z}_1  \mbr{z}_{T-1}  \mbr{z}_T$
%Layer $1$. Layer $T$
%$\mbr{H},\mbr{A},\mbr{Q},\mbr{W},\mbr{\theta}$
%$\mbr{B}$ $\mbr{\hat{x}}$
  \caption{\small\textbf{MLNet architecture.}
A small number $T$ of ISTA iterations are unrolled into a feed-forward network. Each layer applies a non-linear transformation to the current iterate $\mbr{z}_t$, parametrized by $\mbr{H},\mbr{A},\mbr{Q},\mbr{W}$ and $\mbr{\theta}$. Training these parameters using standard back propagation on a set of representative inputs allows the network to approximate the output of the underlying iterative algorithm with much lower complexity.  }
  \label{netProp}
\end{figure}

An iterative solution of (\ref{problemFormulationWithSaprsity}) typically requires hundreds of iterations to converge. This results in prohibitive complexity and unpredictable input-dependent latency unacceptable in real-time applications.
To overcome this limitation, we follow the approach advocated by \cite{gregor-icml-10} and \cite{sprechmann2012learning}, in which a small number $T$ of ISTA iterations are unrolled into a feed-forward neural network, that subsequently undergoes supervised training on typical inputs for a given cost function $f$.
A single ISTA iteration can be written in the form
\begin{equation}
\footnotesize \mbr{z}_{t+1}\ = \sigma_\mbr{\theta}\left( \mbr{z}_{t} - \mbr{W}\rm{diag}(\rho'(\mbr{Q}\mbr{z}_{t}))\mbr{H}^\mathrm{T}\nabla\ell (\mbr{H}\rho(\mbr{A}\mbr{z}_{t}) | \mbr{B} )\right) ,
\end{equation}
where $\mbr{A}=\mbr{Q}=\mbr{D}$, $\mbr{W}=\eta\mbr{D}^T$ and $\mbr{\theta} = \mu\eta\mbr{1}$.
Each such iteration may be viewed as a single layer of the network parameterized by $\mbr{A},\mbr{Q},\mbr{W} \mbox{ and } \mbr{\theta}$, accepting $\mbr{z}_{t}$ as input and producing $\mbr{z}_{t+1}$ as output. Figure \ref{netProp} depicts the network architecture, henceforth referred to as MLNet.

Network supervised training is done by initializing the parameters as prescribed by the ISTA iteration and then adapting them using a stochastic gradient approach which minimizes the reconstruction error $\mathcal{F}$ of the entire network. We use the following empirical loss %over a set of $N$ training examples
\vspace{-1 mm}
\begin{equation}
\label{general_loss}
\mathcal{F} = \frac{1}{N}\sum_{n=1}^{N}f(\mbr{x}_n^*,\hat{\mbr{z}}_T(\mbr{B}_n),\mbr{D}) ,
\end{equation}
which for a large enough training set, $N$, approximates the expected value of $f$ with respect to the distribution of the ground truth signals $\mbr{x}_n^*$. Here, $\hat{\mbr{z}}_T(\mbr{B}_n)$ denotes the output of the network with T layers given the binary images $\mbr{B}_n$ produced from $\mbr{x}_n^\ast$ as the input.

Similarly to \cite{gregor-icml-10} the output of the network and the derivative of the loss with respect to the network parameters are calculated using the forward and back propagation, summarized as Algorithms \ref{algo_front_prop} and \ref{algo_back_prop}, respectively. Practice shows that the training process allows to reduce the number of iterations required by about two orders of magnitude while achieving a comparable reconstruction quality (see Figure \ref{NN_complexity}). To the best of our knowledge, this is the first time a similar strategy is applied to reconstruction problems with a non-Euclidean data fitting term.

\begin{algorithm}[h!]

 \KwIn{Number of layers $T$,$\mbr{\theta}$,$\mbr{Q}$,$\mbr{D}$,$\mbr{W}$,$\mbr{A}$}
 \KwOut{Reconstructed image $\mbr{\hat{x}}$, \\ \hspace{13 mm}auxiliary variables $\{\mbr{z_t}\}_{t=0}^T$,$\{\mbr{b_t}\}_{t=1}^T$ }

initialize $\mbr{z}_0=0$

 \For{t = 1,2,\dots,T}{
 $\footnotesize{\mbr{b}_t = \mbr{z}_{t-1} - \mbr{W}\mr{diag}(\rho'(\mbr{Q}\mbr{z}_{t-1}))\mbr{H}^\mathrm{T}\nabla \ell (\mbr{H}\rho(\mbr{A}\mbr{z}_{t-1}))}$\\
 $\footnotesize{\mbr{z}_t = \sigma_\mbr{\theta}(\mbr{b}_t)}$
 }
$\footnotesize{\mbr{\hat{x}}=\rho(\mbr{D}\mbr{z}_T}$)
\vspace{2 mm}
 \caption{Forward propagation through MLNet.}
\label{algo_front_prop}
\end{algorithm}

\begin{algorithm}[h!]

 \KwIn{Loss $\mathcal{F}$, outputs of \ref{algo_front_prop}: $\{\mbr{z_t}\}_{t=0}^T$,$\{\mbr{b_t}\}_{t=1}^T$}
 \KwOut{Gradients of the loss w.r.t. network parameters $\footnotesize{\mbr{\delta W}}$,$\footnotesize{\mbr{\delta A}}$,$\footnotesize{\mbr{\delta Q}}$,$\footnotesize{\mbr{\delta \theta}}$}
 \vspace{1 mm}
 initialize $\footnotesize{\mbr{\delta W}^T=\mbr{\delta A} = \mbr{\delta Q}=\mbr{0}}$,$\footnotesize{\mbr{\delta\theta}=\mbr{0}}$,$\footnotesize{\mbr{\delta z}_T = \frac{d \mathcal{F}}{d \mbr{z}_T}} $\\

 \For{t = T,T-1,\dots,1}{

% $ \footnotesize{\mbr{a} = \mbr{Az}_{t-1}}$ \\
% $ \footnotesize{\mbr{b} = \mbr{Qz}_{t-1}}$ \\
% $ \footnotesize{\mbr{c} = \mbr{Az}_{t}}$ \\
% $ \footnotesize{\mbr{d} = \mbr{Qz}_{t}}$ \\
% $ \footnotesize{\mbr{e} = \mbr{H} \mr{diag}(\rho'(\mbr{b}))}$ \\
%
% $ \footnotesize{\mbr{\delta b} = \mbr{\delta z}_t \cdot \mr{diag}(\sigma_\mbr{\theta}'(\mbr{b}_t)) } $\\
%
% $ \footnotesize{\mbr{\delta W} = \mbr{\delta W} - \mbr{\delta b} \nabla\ell(\mbr{H}\rho(\mbr{a}))^T \mbr{e} } $ \\
%
% $ \footnotesize{\mbr{\delta A} = \mbr{\delta A} - \mr{diag}(\rho'(\mbr{a}))\mbr{H}^T \nabla^2 \ell(\mbr{H}\rho(\mbr{a}))^T \mbr{e} \mbr{W}^T \mbr{\delta b}_t \mbr{z}_{t-1}^T  }$\\
%
% $ \footnotesize{\mbr{\delta Q} = \mbr{\delta Q} - \mr{diag}(\mbr{H}^T \nabla\ell(\mbr{H} \rho(\mbr{a})))\mr{diag}(\rho''(\mbr{b}))\mbr{W}^T \mbr{\delta b} \mbr{z}_{t-1}^T}$\\
%
% $ \footnotesize{\mbr{\delta \theta} = \mbr{\delta z} \frac{\partial \sigma_\mbr{\theta} (\mbr{b}_t)}{\partial \mbr{\theta}}}$ \\
%
% $ \footnotesize{ \mbr{F} = \mbr{W} \mr{diag}(\rho'(\mbr{d})) \mbr{H}^T  \nabla^2\ell(\mbr{H}\rho(\mbr{c}) \mbr{H} \mr{diag}(\rho'(\mbr{c})\mbr{A}))  }$ \\
% $ \footnotesize{ \mbr{G} = \nabla\ell(\mbr{H}\rho(\mbr{c})^T\mbr{H} \mr{diag}(\rho''(\mbr{d})) \mr{diag}(\mbr{W}^T \mbr{\delta b}^T)\mbr{Q}  }$ \\
%
% $ \footnotesize{\mbr{\delta z}_{t-1} = \mbr{\delta b}^T ( \mbr{I} - \mbr{F} ) - \mbr{G}      }   $ \\

 $ \footnotesize{\mbr{a}^{(1)} = \mbr{Az}_{t-1}}$ \\
 $ \footnotesize{\mbr{a}^{(2)} = \mbr{Qz}_{t-1}}$ \\
 $ \footnotesize{\mbr{a}^{(3)} = \mbr{Az}_{t}}$ \\
 $ \footnotesize{\mbr{a}^{(4)} = \mbr{Qz}_{t}}$ \\
 $ \footnotesize{\mbr{a}^{(5)} = \mbr{H} \mr{diag}(\rho'(\mbr{a}^{(2)}))}$ \\

 $ \footnotesize{\mbr{\delta b} = \mbr{\delta z}_t \mr{diag}(\sigma_\mbr{\theta}'(\mbr{b}_t)) } $\\

 $ \footnotesize{\mbr{\delta W} = \mbr{\delta W} - \mbr{\delta b} \nabla\ell(\mbr{H}\rho(\mbr{a}^{(1)}))^T \mbr{a}^{(5)} } $ \\

 $ \footnotesize{\mbr{\delta A} = \mbr{\delta A} - \mr{diag}(\rho'(\mbr{a}^{(1)}))\mbr{H}^T \nabla^2 \ell(\mbr{H}\rho(\mbr{a}^{(1)}))^T \mbr{a}^{(5)} \mbr{W}^T \mbr{\delta b}_t \mbr{z}_{t-1}^T  }$\\

 $ \footnotesize{\mbr{\delta Q} = \mbr{\delta Q} - \mr{diag}(\mbr{H}^T \nabla\ell(\mbr{H} \rho(\mbr{a}^{(1)})))\mr{diag}(\rho''(\mbr{a}^{(2)}))\mbr{W}^T \mbr{\delta b} \mbr{z}_{t-1}^T}$\\

 $ \footnotesize{\mbr{\delta \theta} = \mbr{\delta \theta} - \mbr{\delta z} \frac{\partial \sigma_\mbr{\theta} (\mbr{b}_t)}{\partial \mbr{\theta}}}$ \\

 $ \footnotesize{ \mbr{F} = \mbr{W} \mr{diag}(\rho'(\mbr{a}^{(4)})) \mbr{H}^T  \nabla^2\ell(\mbr{H}\rho(\mbr{a}^{(3)}) \mbr{H} \mr{diag}(\rho'(\mbr{a}^{(3)})\mbr{A}))  }$ \\
 $ \footnotesize{ \mbr{G} = \nabla\ell(\mbr{H}\rho(\mbr{a}^{(3)})^T\mbr{H} \mr{diag}(\rho''(\mbr{a}^{(4)})) \mr{diag}(\mbr{W}^T \mbr{\delta b}^T)\mbr{Q}  }$ \\

 $ \footnotesize{\mbr{\delta z}_{t-1} = \mbr{\delta b}^T ( \mbr{I} - \mbr{F} ) - \mbr{G}      }   $ \\

 }
\vspace{2 mm}
 \caption{Gradient computation by back propagation through MLNet. We follow the shorthand notation common in the machine learning literature, denoting by $\delta*$ the gradient of the scalar loss $\mathcal{F}$ w.r.t. the network parameter $*$. Note that $\mbr{\delta \mbr{D}}$ is calculated separately as it only depends on the last iteration of the network.}
\label{algo_back_prop}

\end{algorithm}

\section{Results}
\label{sec_results}
\subsection{Low light} \label{lowlight}
This experiment emphasizes the superiority of MLNet over an unregularized ML reconstruction in low light conditions. Furthermore, it illustrates the tradeoff between reconstruction quality and computational complexity.

%During the course of the experiment we took a low-resolution image, generated a high-resolution binary image simulating the %sensor's acquisition process and tested different reconstruction methods, comparing their results and computational complexity.

As a low-resolution ground truth we used the Lena gray-level image normalized to the low light range $[0,10]$. A high-resolution binary image was created in the following way: First, a photon flux average rate image was created by up-sampling and low-pass filtering the ground truth image using a $2D$ Gaussian filter with a standard deviation of $3$ and an up-sampling factor of $5$. Then, an acquisition process was simulated as a realization of the Poisson probability distribution of the given average rate image followed by binarization using a fixed $5\times 5$ threshold pattern consisting of the values $q_i\in\{1,...,10\}$ distributed uniformly and tiled over the entire image.

An image was reconstructed from the binary measurements using the unregularized ML and using MLNet with several depths $T$ operating on $8 \times 8$ overlapping patches. The function $\rho$ was set according to \ref{rho} with c=10.

MLNet was trained using stochastic gradient decent optimization on a disjoint set of $30,000$ $8 \times 8$ image patches normalized to the same range. In each iteration the descent direction was calculated on a random minibatch of $1,000$, low-resolution ground truth patches and their corresponding randomly generated high-resolution binary measurements. The optimization over the parameters was done via round-robin \textit{i.e.} the parameters were optimized sequentially in a circular manner. A validation set was used to prevent over-fitting.
The loss objective $f$ minimized during the training process was the standard squared error,
\begin{equation}
f = \frac{1}{2}\| \mbr{x}_n^* - \rho(\mbr{D} \mbr{z}_T(\mbr{B}_n)) \|_2^2,
\end{equation}

% $L(\lambda,\lambda^*)=||\lambda-\lambda^*||_2^2$.
\noindent where $\mbr{x}_n^*$ denote the training ground truth patches.

\begin{figure}
%\vspace{-40pt}
    \centering
        \begin{tabular}{ c c }
            \includegraphics[width = 0.2\textwidth]{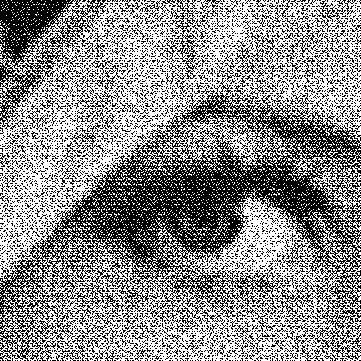} &
            \includegraphics[width = 0.2\textwidth]{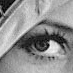} \\
            (a) Binary image & (b) Ground Truth \\
            \includegraphics[width = 0.2\textwidth]{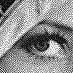} &
            \includegraphics[width = 0.2\textwidth]{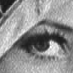} \\
            (c) Unregularized ML  & (d) MLNet \\
          \end{tabular}   \\ % \vspace{6pt}
\vspace{2 mm}
    \caption{\small \textbf{Low light reconstruction.}
        Lena's image was normalized to $[0,10]$ from which four input binary images were emulated.
        Depicted is a zoomed in fragment of the images: (a) input binary image; (b) low-resolution ground truth; (c) unregularized ML reconstruction (PSNR=$16.6$ \rm{dB}); (d) MLNet reconstruction (PSNR=$27.5$ \rm{dB}). MLNet performs better both quantitatively and qualitatively. The reader in encouraged to zoom-in on the images.
}
    \label{Lena}
\end{figure}

Figure \ref{Lena} compares the unregularized ML reconstruction to the reconstruction of MLNet with $T=8$ layers without patch overlap. As can be seen MLNet significantly outperforms the reconstruction obtained by the unregularized ML reaching $27.5$ dB PSNR compared to only $16.6$ dB.

\begin{figure}[]

\includegraphics[width = 1\columnwidth]{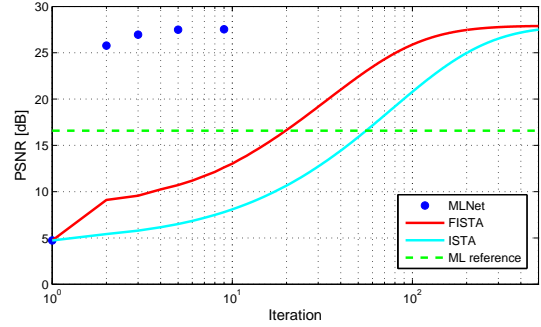}
\caption{\small \textbf{Reconstruction accuracy under budgeted computational complexity.} The plot shows the reconstruction quality for the iterative ISTA and FISTA solvers with the ML data fitting term when terminated after a given number of iterations, and the proposed MLNet with an equivalent number of layers. As a reference, the performance of the unregularized ML is shown. Iteration $1$ represents the initialization for all algorithms.}
\label{NN_complexity}
\end{figure}

In order to evaluate the tradeoff between reconstruction quality and computational complexity, in Figure \ref{NN_complexity} we compared the performance of MLNet with different depths with the reconstruction quality achieved by ISTA and FISTA terminated after a comparable number of iterations (thus, the compared reconstruction algorithms have approximately the same complexity). As a reference,  we also show the unregularized ML reconstruction. We observe that MLNet produces acceptable reconstruction quality already for $T=4$ layers, which takes FISTA around $200$ iteration (at least $50$ times higher complexity) to achieve. Similar speedup was observed consistently in other experiments not reported here due to lack of space.

\subsection{Multiple exposures} \label{multipleExposures}
\begin{figure}[]

\includegraphics[width = 1\columnwidth]{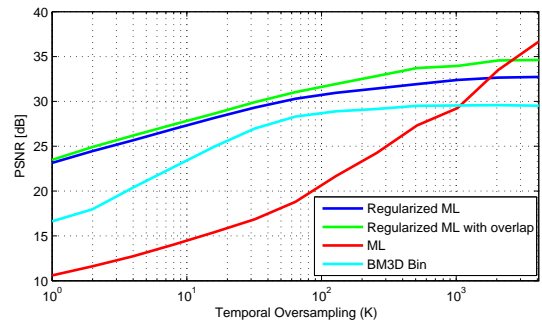}
\caption{\small \textbf{Reconstruction quality for different amounts of temporal oversampling $K$.} The plot shows the effect of the number of exposures $K$ over the reconstruction quality for the unregularized and regularized ML estimators (the latter solved using FISTA) as well as the reconstruction using BM3D with binning.}

%It can be seen that the sparse prior allows a more accurate reconstruction from less data.}

\label{MultipleExpGraph}
\vspace{-1em}
\end{figure}

In this experiment, we reconstructed a synthetic static low light scene using a sequence of multiple exposures. As is often the case in estimation problems, as the number of measurements increases the effect of regularization slightly degrades the image quality compared to the unregularized estimator. Figure \ref{MultipleExpGraph}, summarizing the experiment described in this section, shows that the number of measurements required by the unregularized ML estimation to outperform reconstruction with a sparse prior is unacceptably high for most practical purposes. In addition, it is evident that a BM3D (\cite{makitalooptimal},\cite{makitalo2009inversion}) based method outperforms the ML estimator but is inferior to its regularized version.

The scene used for the experiment was a low light image with gray values normalized to the range $[0,10]$. A high resolution average rate image was created as in Section \ref{lowlight} by up-sampling the ground truth image by a factor of $3$, followed by a Gaussian filter with a standard deviation of $1.5$. $K$ different binary images were created by generating independent photon count images via a realization of the Poisson process followed by binarization. Binarization was performed using a threshold pattern of size $3 \times 3$ containing the threshold values $q_i\in\{1,...,9\}$  tiled regularly over the entire image. $c=10$ was selected for function (\ref{rho}). FISTA was run until convergence with the regularization weight $\mu = 4$ and a backtracking procedure to determine the step size $\eta$ at each iteration.
Note that the data fitting term (\ref{likelihood}) increases linearly with $K$, therefore, in practice keeping $\mu$ constant reduces the regularization weight as $K$ increases.

As a comparison to Poisson denoising algorithms we reconstruct the image using BM3D. As each pixels value is a Bernoulli variable summing over $K$ exposures yields a Binomial variable that can be approximated as a Poisson distribution. This image can be denoised using the Poisson adapted version of BM3D. The underlying photon flux rate $\lambda_j$ can be inferred from $p_j$ and $q_j$, then spatial binning can be applied according to the spatial oversampling of the binary sensor.

\begin{figure}
%\vspace{-40pt}
    \centering
        \begin{tabular}{ c c c}
            \hspace{-2 mm}\includegraphics[width = 0.15\textwidth]{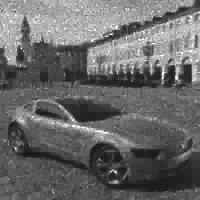} & \hspace{-4 mm}
            \includegraphics[width = 0.15\textwidth]{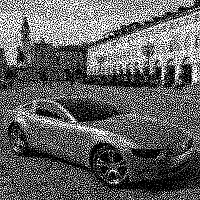} & \hspace{-3 mm}
            \includegraphics[width = 0.15\textwidth]{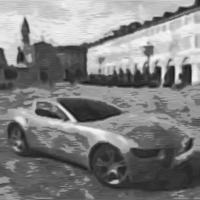} \\
            \hspace{-2 mm}(a) \small{Reg ML, $K=1$}  & \hspace{-3 mm} (b) \small{ML, $K=1$} & \hspace{-4 mm} (c) \small{BM3D, $K=1$}\\

            \hspace{-2 mm}\includegraphics[width = 0.15\textwidth]{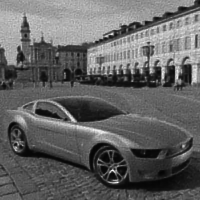} & \hspace{-4 mm}
            \includegraphics[width = 0.15\textwidth]{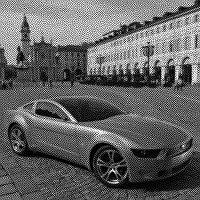} & \hspace{-3 mm}
            \includegraphics[width = 0.15\textwidth]{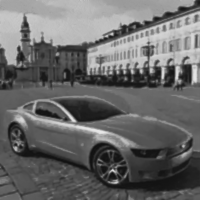} \\
            \hspace{-2 mm}(d) \small{Reg ML, $K=32$}  & \hspace{-3 mm} (e) \small{ML, $K=32$} & \hspace{-4 mm} (f) \small{BM3D, $K=32$}\\
            \end{tabular}   \\ % \vspace{6pt}
          \vspace{2 mm}
    \caption{\small \textbf{Reconstruction from multiple exposures.}
        (a), (b) and (c) are the reconstructions of Regularized ML, ML and BM3D with binning using a single binary image ($K=1$);
        (d), (e) and (f) are the reconstructions of Regularized ML, ML and BM3D with binning using ($K=32$) binary images.
        %(a) Regularized ML image reconstruction from a single ($K=1$) binary image (PSNR=$23.2$ \rm{dB});
        %(b) ML reconstruction from a single binary image (PSNR=$10.6$ \rm{dB});
        %(c) \textcolor{red}{Modified BM3D with binning from a single ($K=1$) binary image (PSNR=$16.6$ \rm{dB})};
        %(d) Regularized ML reconstruction from $K=32$ binary images (PSNR=$29.3$ \rm{dB});
        %(e) ML reconstruction from $K=32$ binary images (PSNR=$16.9$ \rm{dB}). Non overlapping patches were used.
        %(f) \textcolor{red}{Modified BM3D with binning from $K=32$ binary images (PSNR=$27$ \rm{dB})};
}
    \label{MultipleExpImages}
    \vspace{-1em}
\end{figure}

Figure \ref{MultipleExpGraph} shows the reconstruction PSNR of BM3D, regularized ML solved via FISTA, with and without overlapping patches, and its unregularized counterpart for different values of the temporal oversampling $K$. In cases where overlap was used the patch shift was four pixels and the overlapping patches were averaged. Examples of reconstructed images for $K = 1$ and $K=32$ are presented in Figure \ref{MultipleExpImages}.

It is evident from the figures that the sparse prior enables a significant improvement in reconstruction quality for a vast range of temporal oversampling values. While the regularized solution reaches a PSNR value of $23.2$ \rm{dB} using a single exposure, its unregularized counterpart reaches similar reconstruction accuracy only after about $200$ exposures. Using $K=32$ exposures, regularized ML achieves $29.9$ \rm{dB} PSNR and the reconstructed image looks visually satisfying while the unregularized ML reconstruction looks poor yielding only $16.9$ \rm{dB} PSNR proving the superiority of a sparse reconstruction for practical reasons. It is worth mentioning that the intersection of reconstruction quality with and without a sparse prior is at around $2000$ exposures which is well beyond any practical limit while assuming a static scene. Furthermore, we can conclude that taking into account the correct statistics of the problem results in a better reconstruction compared to a general Poisson denoising technique.


\subsection{HDR imaging emulation} \label{HDR}
This experiment applies our method to HDR images while demonstrating the feasibility of real-time image reconstruction from binary measurements using MLNet, with inputs emulated using a DSLR sensor.

\begin{figure*}[h]
    \centering
        \begin{tabular}{ c c c c }
           \includegraphics[width = 0.23\textwidth]{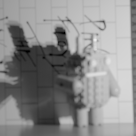} &
           \includegraphics[width = 0.23\textwidth]{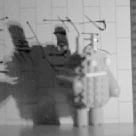} &
           \includegraphics[width = 0.23\textwidth]{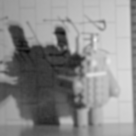} &
           \includegraphics[width = 0.23\textwidth]{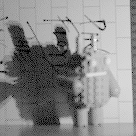} \\

           \includegraphics[width = 0.23\textwidth]{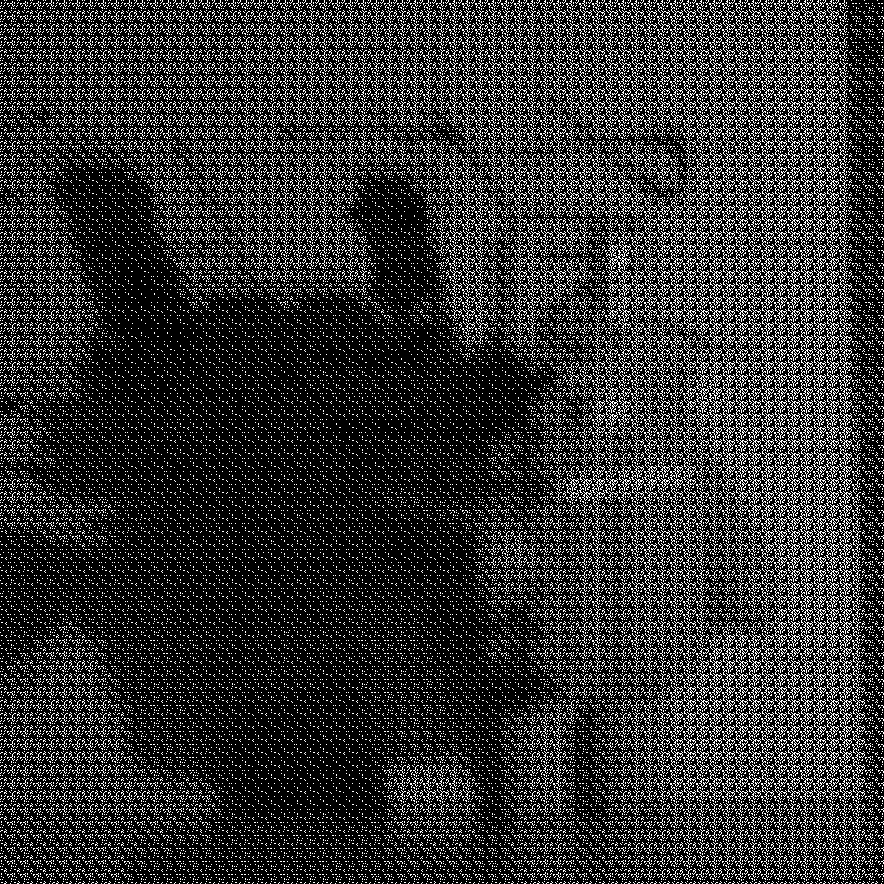} &
           \includegraphics[width = 0.23\textwidth]{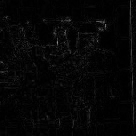} &
           \includegraphics[width = 0.23\textwidth]{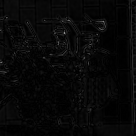} &
           \includegraphics[width = 0.23\textwidth]{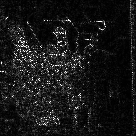} \\
            & PSNR=$41$ \rm{dB} & PSNR=$40$ \rm{dB} & PSNR=$24.4$ \rm{dB} \\
            & & & \\

           \includegraphics[width = 0.23\textwidth]{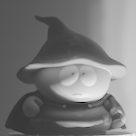} &
           \includegraphics[width = 0.23\textwidth]{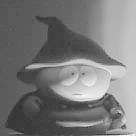} &
           \includegraphics[width = 0.23\textwidth]{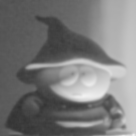} &
           \includegraphics[width = 0.23\textwidth]{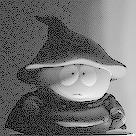} \\

           \includegraphics[width = 0.23\textwidth]{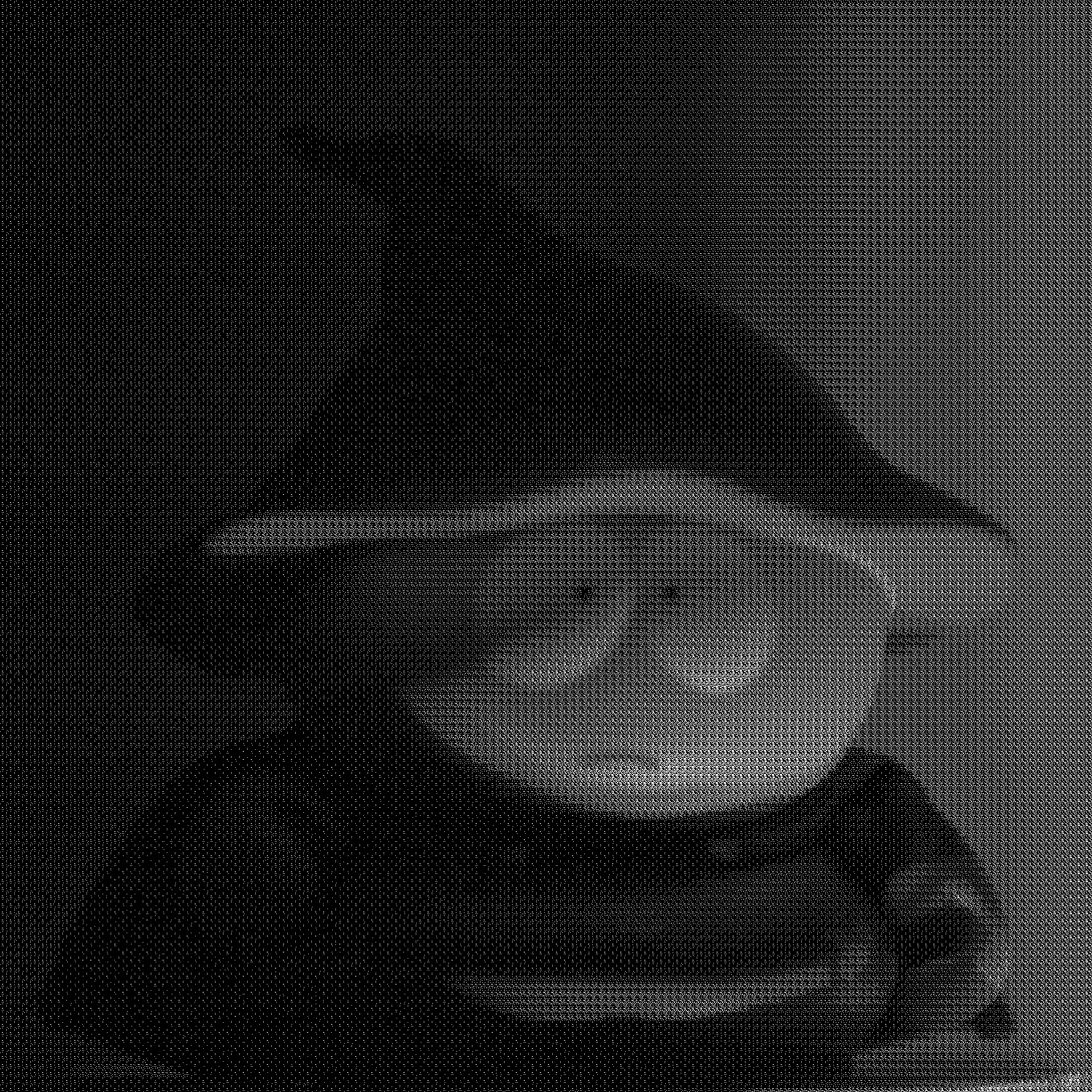} &
           \includegraphics[width = 0.23\textwidth]{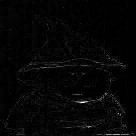} &
           \includegraphics[width = 0.23\textwidth]{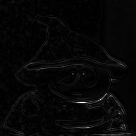} &
           \includegraphics[width = 0.23\textwidth]{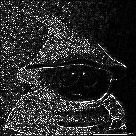} \\
            & PSNR=$35.6$ \rm{dB} & PSNR=$35.5$ \rm{dB} & PSNR=$17.9$ \rm{dB} \\
           (a) & (b) & (c) & (d) \\

          \end{tabular}   \\
          \vspace{2 mm}
    \caption{\small \textbf{High dynamic range reconstruction} of the Lego Android (top rows) and Cartman (bottom rows) images. Column (a) shows the ground truth low resolution image (first and third rows) and the high resolution binary input (second and fourth rows). Column (b) presents regularized ML reconstruction via FISTA with no patch overlap and the difference image with the ground truth. Columns (c) and (d) present the same for MLNet and the unregularized ML algorithms respectively. Note that all images are presented on a logarithmic scale on which the PSNR is calculated. It is evident that the sparse prior enables much better reconstructions of real HDR scenes captured using a DSLR camera. We encourage the reader to zoom-in on the images.}
    \label{hdr}
\end{figure*}

We present the performance of MLNet on real captured scenes with a five orders of magnitude dynamic range. In the absence of an actual sub diffraction limit binary sensor, we emulated its functionality using a Canon DSLR sensor. The spatial oversampling effect was simulated by defocusing the captured frames (the amount of defocusing was set to fit the desired spatial oversampling of $13 \times 13$), thus resulting in a reconstructed image that is smaller by the upsampling factor compared to the native camera resolution. We found that setting the ISO value to $200$ results in a gain of one gray level per photoelectron for pixels belonging to the \rm{G1} Bayer channel, thus allowing direct photoelectrons counting from the sensor raw images (after subtracting the dark value). As the dynamic range of a single RAW image is about four orders of magnitude, in order to capture the entire dynamic range of the scene, a sequence of images with multiple exposure times was taken, starting with the longest exposure and replacing only the saturated content using information from shorter exposures. In order to preserve the underlying Poisson statistics generated by the source light field, we used the additivity property of independent Poisson variables taking exposure times ranging between $0.01$ and $0.2$ seconds to produce an input image of the photoelectron flux with a total dynamic range of five orders of magnitude. A low resolution ground truth image was created by averaging $10$ such photoelectron count images and down-sampling, as shown in Figure \ref{hdr}(a). The scenes presented were set in a dark room and were lighten with a LED flash light producing a dynamic range $[0,10^5]$.

The binary image was generated using a threshold pattern selected according to the following rationale: Due to the fact that the variance of a Poisson distribution is equal to its average rate, a single binary pixel would show informative behavior in the interval $\left[q-2\sqrt{q},q+2\sqrt{q}\right]$. The number of threshold values required in order to cover the range $[0,10^5]$ with such intervals is approximately $155$, explaining the choice of the spatial oversampling ratio to be $13 \times 13$. The remaining threshold values were spread over the lower region of the dynamic range using an analogous technique. The resulting binary image is depicted in Figure \ref{hdr}(second and fourth rows of the column (a)).

The dictionary used for the regularization term in the regularized ML estimator and the initialization of MLNet was a $k$SVD dictionary with $257$ atoms of size $8 \times 8$ that was trained on the \rm{log} intensity of patches from the Stanford HDR dataset \cite{xiao2002high}.
 A regularization weight of $\mu=50$ was empirically chosen for FISTA using a small validation set from \cite{xiao2002high}. In order to reduce the number of FISTA iterations, we altered Algorithm \ref{algo_fista} by resetting the step size $\eta$ to its initial value every five iterations. Our experience shows that this allows faster convergence with our non-Euclidean data fitting terms, as it allows longer steps on sub-domains where the Lipschitz constant of the gradient of $\ell$ is smaller. BM3D was used as explained in section \ref{multipleExposures}.

As in the low light experiment detailed in Section \ref{lowlight}, MLNet was trained using data similar to the data used for the $k$SVD dictionary training process. We trained MLNet to have $T=25$ recurrent layers with $c=10^5$ using a stochastic gradient optimization, similar to the one described in Section \ref{lowlight}, only using a smaller minibatch size due to memory limitations. In this experiment we used an objective function of the form
\vspace{-1 mm}
\begin{equation}\label{loss_hdr}
f = \frac{1}{2}\| \rm{log}(1+\mbr{x}_n^*) - \rm{log}(1+\rho(\mbr{D} \mbr{z}_T(\mbr{B}_n))) \|_2^2
\end{equation}

\noindent which we found to be a more meaningful error criterion for HDR images. Reconstruction results using FISTA and MLNet are presented in Figure \ref{hdr}. FISTA was run without patch overlap and MLNet was run with full overlap, followed by patch averaging. Due to the images being HDR we chose to present them in a \rm{log} scale. In addition, difference images between the \rm{log} of the ground truth images and the \rm{log} of the reconstructions are also presented. Interestingly, these results show that training on synthetic data from Stanford HDR dataset results in a neural net that performs well on real world data captured by a DSLR sensor.

Reconstruction time required per algorithm for Figure \ref{hdr}(third row, column (a)) are presented in Table \ref{recon_time_table} showing that MLNet with $T=25$ recurrent layers offers a speedup between three to four orders of magnitude compared to the regularized ML reconstruction via FISTA. Note that the timing reported for MLNet in the seconds range was obtained on a regular Intel Core i7 CPU. MLNet is amenable to heavy parallelization (at least, at the patch level) and is hardware-friendly because of its regular fixed-latency data flow. We are currently developing an FPGA prototype demonstrating a video-rate real time reconstruction. It is evident that Poisson denoising algorithms such as BM3D fail once the number of exposures is low and the scene is of high dynamic range. Due to lack of space and poor quality we chose not to show these reconstructions in Figure \ref{hdr}.

Although the images reconstructed by MLNet offer high PSNR values they seem to be slightly blurred. While we currently do not fully understand the source of this blur and whether it can be mitigated by more sophisticated network architectures, we believe that MLNet offers an interesting tradeoff between reconstruction quality and computational complexity for real world applications and HDR imaging from a single binary image.

\begin{table}
\begin{tabular}{| l | c | c | c | c |}
        \hline
        Algorithm & BM3D & ML & FISTA & MLNet \\
        \hline
        Time & 20 min & $6.2$ hrs & $2.8$ hrs & $4$ sec \\
        PSNR [\rm{dB}] & 6.7 & $17.9$ & $35.6$ & $35.5$ \\
        \hline
\end{tabular}
\vspace{2 mm}
\caption{\small \textbf{HDR image reconstruction time comparison}. A comparison of the Cartman image reconstruction time. FISTA and MLNet were both run without patch overlap.}
\label{recon_time_table}
\vspace{-1em}
\end{table} 

\section{Conclusions and future work}
\label{sec_conclusions}
In this work we discussed image reconstruction from a set of dense binary measurements, motivated by recent advances in the sensing hardware. We demonstrated the superiority of the proposed regularized ML reconstruction over "vanilla ML" as well as Poisson denoising algorithms such as BM3D, showing that the regularized version achieves similar performance as its unregularized counterpart with about three orders of magnitude less measurements. Taking computational complexity into account, we also showed how to compute an efficient and hardware-friendly approximation to the reconstruction algorithm. Promising results were shown on synthetic data as well as on data emulated using multiple exposures of a regular CMOS sensor.
It is worthwhile noting that the training loss function (\ref{general_loss}) allows great freedom of choice, including fully unsupervised or semi-supervised training regimes. While we clearly observed non-negligible effects of such a choice on the reconstruction quality, in this work we opted for the standard MSE criterion, deferring the detailed study of the loss functions to future work.
We would like to state that the proposed techniques have the same limitations of approaches based on a universal dictionary as opposed to a prior trained from the data themselves. They also suffer from known shortcomings of patch-based approaches that average reconstructed overlapping patches: while each of the patches admits the prior, their average might not. In future work, we intend to explore ways of overcoming these limitations.

\clearpage
{\small
\bibliographystyle{ieee}
\bibliography{MLNet_V12}
}

%\section{Appendix}
%\input{appendix_V4.tex}

\end{document}